\documentclass{article}     % onecolumn (ditto)
\pdfoutput=1 
\usepackage{graphicx}
\usepackage{latexsym}
\usepackage{natbib}
\usepackage{amsmath}
\usepackage{amsfonts}
\usepackage{amssymb}
\usepackage{url}
\usepackage[pdftex]{hyperref}

\usepackage{bbm}

\newcommand{\dataset}{{\cal D}}

\newtheorem{theorem}{Theorem}

\newtheorem{problem}{Problem}
\newtheorem{conjecture}[theorem]{Conjecture}
\newcommand{\Hypo}{\mathcal{H}}
\newcommand{\EHypo}{\widehat{\mathcal{H}}}

\newcommand{\X}{\mathcal{X}}

\newcommand{\D}{\mathcal{D}}
\newcommand{\R}{\mathbb{R}}

\newcommand{\Ecal}{\mathcal{E}}
\newcommand{\Ehat}{\widehat{\mathcal{E}}}

\newcommand{\Uhat}{\widehat{\mathcal{U}}}

\newcommand{\err}{\mathrm{err}}

\newcommand{\indicator}[1]{\mathbbm{1}{\left[ {#1} \right] }}
\newcommand{\E}[1]{\mathbf{E}\left[ #1\right] }

\DeclareMathOperator*{\argmin}{arg\,min}

\date{}

\title{Some Open Problems in Optimal
  AdaBoost and Decision Stumps}
\usepackage{times}
\author{Joshua Belanich\\ 
Department of Computer Science\\
Stanford University\\
Stanford, CA USA\\
\url{jbelanich@cs.stanford.edu}\\
 \and
 Luis E. Ortiz\\
 Department of Computer Science\\
 Stony Brook University\\
Stony Brook, NY USA\\
\url{leortiz@cs.stonybrook.edu}
 }

\begin{document}

\maketitle

\begin{abstract}
% This is a great paper and it has a concise abstract. 
  The significance of the study of the theoretical and practical
  properties of AdaBoost is unquestionable, given its simplicity, wide
  practical use, and effectiveness on real-world datasets. Here we
  present a few open problems regarding the behavior of ``Optimal
  AdaBoost,'' a term coined by Rudin, Daubechies, and Schapire in 2004
  to label the simple version of the standard AdaBoost algorithm in
  which the weak learner that AdaBoost uses always outputs
  the weak classifier with lowest weighted error among the respective
  hypothesis class of weak classifiers implicit in the weak learner.
  We concentrate on the standard, ``vanilla'' version of Optimal
  AdaBoost for binary classification that results from using an
  exponential-loss upper bound on the misclassification training
  error. We present two types of open problems. One deals with general
  weak hypotheses. The other deals with the particular case of
  decision stumps, as often and commonly used in practice. Answers to
  the open problems can have immediate significant impact to
  (1) cementing previously established results on asymptotic convergence properties of Optimal AdaBoost, for finite datasets,
  which in turn can be the start to any convergence-rate analysis; (2)
  understanding the weak-hypotheses class of effective decision stumps generated
  from data, which we have empirically observed to be significantly smaller than
  the typically obtained class, as well as the
  effect on the weak learner's running time and previously established
  improved bounds on the generalization performance of Optimal AdaBoost
  classifiers; and (3) shedding some light on the ``self control'' that
  AdaBoost tends to exhibit in practice.
\end{abstract}

% \begin{keywords}
% optimal AdaBoost; asymptotic convergence; decision stumps
% \end{keywords}

\section{Introduction}

Due to space constraints, we concentrate on stating the open problems
and conjectures without entering into the details. We refer the
reader to our manuscript on the convergence properties of Optimal
AdaBoost for additional details~\citep{belanich12}, recently updated
for presentation and clarification purposes. We also refer the reader
to that manuscript for further discussion of the important
implications, briefly listed in the
Abstract, that answers to the
open problems and conjectures stated here would have.

We note that our main interest is not highly synthetic, ``low-dimensional'' examples that contradict the
conjectures unless, of course, such examples are the simple start of
more sophisticated
constructions of non-trivial and realistic counterexamples.

\paragraph{Technical Preliminaries and Notation.} Let $\X$ denote the \emph{feature space} (i.e., the set of all inputs) and 
%$\Y \equiv \{-1,1\}$ 
$\{-1,+1\}$ be the set of \emph{(binary) output labels}. To simplify notation, let 
%$\D \equiv \X \times \Y$ 
$\D \equiv \X \times \{-1,+1\}$ 
be the set of possible input-output pairs. In typical AdaBoost, we
want to learn from a given, \emph{fixed} dataset of $m$ training
examples $D \equiv
\{(x^{(1)},y^{(1)}),(x^{(2)},y^{(2)}),\ldots,(x^{(m)},y^{(m)})\}$,
where each input-output pair $(x^{(l)},y^{(l)}) \in \D$, for all
examples $l = 1,\ldots,m$.
We make the standard assumption that each example 
in $D$ comes i.i.d. from a \emph{probability space} $(\D,\Sigma,P)$, where
$\D$ is the \emph{outcome space}, $\Sigma$ is the
\emph{($\sigma$-algebra) set of possible events} with respect to $\D$
(i.e., subsets of $\D$), and $P : \Sigma \to \R$ is the \emph{probability measure}.

We denote \emph{the set of hypotheses that the weak learner that
AdaBoost uses}, or simply \emph{the weak-hypothesis class}, by
$\Hypo$.
%, often referred to within the boosting
%context as the \emph{weak-hypothesis class}, and its elements as
%\emph{potentially selected weak hypotheses}.
We say that 
%the weak-hypothesis class 
$\Hypo$ 
%that the Optimal AdaBoost weak-learner uses 
is \emph{AdaBoost-natural} with respect to $D$ 
%a given dataset $D$ 
if (1) the hypothesis $h$, such that, for all $x \in \X$,
$h(x)=1$, is in $\Hypo$; (2) if $h \in \Hypo$, then $-h \in \Hypo$;
%, where by $-h$ we mean the function $h'(x) \equiv - h(x)$; 
and (3) for
all $h \in \Hypo$, there exists an $(x,y) \in D$ such that
$h(x) \neq y$.

In our work, we study Optimal AdaBoost as a \emph{dynamical system of
  the weights $w$ over the examples in $D$}; we also refer to such $w$
as the \emph{example or sample weights}, in a way similar to previous
work~\citep{RudinDynamics}. In particular, we take a dynamical system view to the Optimal AdaBoost update rule of
the example weights $w$ on $D$. That is, each $w$ is a probability
distribution over the $m$ examples in $D$. The set of all such $w$'s, denoted by $\Delta_m$, corresponds
to the \emph{state space of the AdaBoost-induced dynamical system}. Denote by $(w_1,w_2,\ldots)$ the infinite
sequence of examples weights that AdaBoost would generate if it were run
infinitely (i.e., the total number of rounds $T \to \infty$). We
deviate slightly in the initialization of $w_1$, which is often
uniform over the set of examples: i.e., $w_1(l) = \frac{1}{m}$ for
$l=1,\ldots,m$. Instead, we let $w_1 \sim \mathrm{Uniform}(\Delta_m)$.
% to be drawn uniformly from the set of all probability distributions over $m$ events. 

We also assume
that the weak learner has a \emph{deterministic tie-braking rule}; i.e., denoting $\err(h; w, D) \equiv \sum_{l=1}^m w(l) \indicator{h(x^{(l)})
\neq y^{(l)}}$ and $\Hypo(w, D) \equiv \argmin_{h \in \Hypo} \err(h; w,
D)$,  for every example weight $w$ that AdaBoost could
generate, \emph{the weak learner always outputs the same weak
  hypothesis} $h^* \in \Hypo(w, D)$. We call $h^*$ the \emph{(weak-learner's) representative
hypothesis of the set $\Hypo(w, D)$}. In addition, we assume that if the set $\Hypo(w,D)
= \Hypo(w',D)$ for any other $w'\neq w$, then the representative hypothesis of $\Hypo(w',D)$
is also $h^*$.

\iffalse
We implicitly assume throughout that the standard \emph{Weak-Learning Assumption} used
in AdaBoost holds; that is, given any dataset with $m$ examples, there exists a real-value
$\gamma \in (0,1/2)$ such that for all example-weights $w \in \Delta_m$, there exists
an $\eta \in \Ecal$ %the achieved optimal
that achieves weighted error
$\eta \cdot w \leq \frac 12 - \gamma < \frac 12$. 
\fi

\section{The No-Ties Conjecture}

The following conjecture essentially states that Optimal AdaBoost
eventually has no ties in the selection of the best weak-classifier at
each round. We use this no-ties condition to establish the convergence of the AdaBoost
classifier, its generalization error, and in fact, the time/per-round
average of any $L_1$ measurable function of the $w_t$'s generated by
Optimal AdaBoost, which include the output classifier's margin, the
example margins, as well as the
weighted error $\epsilon_t$'s and the weak-hypothesis weight
$\alpha_t$'s of the selected hypotheis $h_t$'s at each round $t$.

We denote by $\mu$ both the Borel and the countable measure, as
appropriate and clear from context. In the statement below, we
assume that the characterization of the set of all $\mu$-probability
spaces, and all $\mu$-measurable spaces over
$\Hypo$, each depend on their own different set of parameters
with Borel or counting measurable spaces, as appropriate for the
corresponding $\sigma$-algebras.
\begin{conjecture}{\bf (No-Ties Conjecture)}
For $\mu$-almost all probability spaces $(\D,\Sigma,P)$ and any dataset
$D \sim P$, and
$\mu$-almost all $\Hypo$ that are AdaBoost-natural with respect to
$D$, there exist $m',T' > 0$, 
%possibly dependent on $P$ and $\Hypo$ 
such that if $m>m'$ is the size of $D$,
then, $P$-almost surely, either (1), for all
$t > T'$ rounds of Optimal AdaBoost,
either (1.a) $|\Hypo(w_t, D)| = 1$; or (1.b)
for all pairs $h_t,h_t' \in \Hypo(w_t,D)$,  $h_t(x^{(l)}) =
h_t'(x^{(l)})$ for all $l=1,\ldots,m$; or (2) $\lim_{t \to
  \infty} \sum_{l=1}^m w_t(l) \indicator{h_t(x^{(l)}) \neq
  h_t'(x^{(l)})} = 0$, where 
$h_t,h_t' \in \Hypo(w_t, D)$. 
\end{conjecture}
%We use the fo The next conjecture typically holds for most other classifiers.
\begin{conjecture}{\bf (Measure-Zero-Decision-Boundary Conjecture)}
For $\mu$-almost all probability spaces $(\D,\Sigma,P)$ and any dataset
$D \sim P$, and
$\mu$-almost all $\Hypo$
%possibly dependent on $D$
that are AdaBoost-natural
with respect to $D$, there exist $m' ,T'> 0$, 
%possibly dependent on $P$ and $\Hypo$, 
such that if $m > m'$,
% is the size of $D$ 
and $T > T'$ is the total number of rounds of Optimal AdaBoost, 
then the decision boundary of the binary classifier that Optimal AdaBoost
outputs after $T$ rounds when given dataset $D$ as input has $P$-measure zero.
\end{conjecture}
In our work we employ
tools from ergodic theory to establish our convergence results. 
%In doing so, 
We provide a non-constructive proof of the existence of a
measure for which the Optimal-AdaBoost update is measure-preserving.
\begin{conjecture}{\bf (Constructive-Proof Conjecture)}
For $\mu$-almost all probability spaces $(\D,\Sigma,P)$ and any dataset
$D\sim P$, and
$\mu$-almost all $\Hypo$ that are AdaBoost-natural
with respect to $D$, there exists $m' > 0$, 
%possibly dependent on $P$ and $\Hypo$, 
such that if $m > m'$,
%is the size of $\dataset$ 
% is the number of rounds of Optimal AdaBoost, 
then there exists a constructive proof of existence of a measure for which 
the Optimal AdaBoost weight update is measure-preserving, $P$-almost surely.
\end{conjecture}

\section{AdaBoosting Decision Stumps}

For simplicity, we concentrate on the feature space $\X = [0,1]^n$,
the $n$-dimensional hypercube, so that $\D =[0,1]^n
\times\{-1,+1\}$. Denote by $\EHypo$ \emph{the
finite set
of decision stumps} on the finite
dataset $D$ induced by using the so-called \emph{midpoint rule}. In this rule, we
project $D$ along each feature dimension $i$ and create a
\emph{decision stump} $h$ based on the midpoint between any pair of
distinct consecutive
examples $x_i^{(l_j)} < x_i^{(l_{j+1})}$ with different labels
$y^{(l_j)} \neq y^{(l_{j+1})}$, such that, denoting the corresponding
midpoint by $x_i' = \frac{x_i^{(l_j)} + x_i^{(l_{j+1})}}{2}$, we can define the decision
stump as 
%$h(x) = 1$, if $x_i > x_i'$; and $=-1$, otherwise. 
$h(x) = \mathrm{sign}(x_i - x_i')$.
\iffalse
Of course, because we assume $\Hypo$
is a AdaBoost-natural hypothesis class with respect to $\dataset$, the negation of that
decision stump $-h \in \Hypo$, as are the constant functions: $h(x) =
+1$ and $h(x)=-1$. 
\fi
We eliminate from $\EHypo$ any \emph{dominated hypothesis}; that is,
we do not need to consider any $h \in \EHypo$, such that there exists
another $h' \in \EHypo$, with the property that $h'(x^{(l)}) \neq
y^{(l)} \implies h(x^{(l)}) \neq y^{(l)}$ for all
$l=1,\ldots,m$. Denote the resulting \emph{effective set} by $\Ehat$. Further, denote
by $\Uhat_T \equiv \bigcup_{t=1}^T \{h_t\}$ \emph{the set of unique decision stumps actually
selected by Optimal AdaBoost from $\Ehat$  after $T$ rounds}. 

\begin{problem}{\bf (Bounding the Number of Effective Stumps)}
Given a measurable space $(\D,\Sigma,P)$, a dataset $D \sim P$,
and the number of rounds of Optimal AdaBoost $T$. Provide non-trivial upper and lower bounds on
$|\EHypo|$, $|\Ehat|$, and  $|\Uhat_T|$.
\end{problem}
Trivial upper and lower bounds are $1 <
|\Uhat_T| \leq |\Ehat| \leq |\EHypo| \leq 2(n (m-1)+1)$.
\begin{conjecture}{\bf (Logarithmic Growth on Unique Stumps)}
For $\mu$-almost all probability spaces $(\D,\Sigma,P)$ and any dataset
$D \sim P$, there exist $m' ,T'>
0$, 
%possibly dependent on $P$ and $\Hypo$, 
such that if $m > m'$ 
%is the size of= $\dataset$ 
and $T > T'$,
% is the number of rounds of Optimal AdaBoost, 
we have $\E{\left. |\Uhat_T| \right| D} \leq (\log T + 1)^c$, for some $c \in [1,2)$,
$P$-almost surely.
\end{conjecture}

\iffalse
% Acknowledgments---Will not appear in anonymized version
\acks{This work was supported in part by an NSF CAREER Award IIS-105454.}
\fi

\bibliographystyle{plainnat}
\bibliography{citation}

\iffalse 

\bibliography{yourbibfile}

\appendix

\section{My Proof of Theorem 1}

This is a boring technical proof.

\section{My Proof of Theorem 2}

This is a complete version of a proof sketched in the main text.
\fi

\end{document}